**People's Water Data: Enabling Reliable Field Data Generation and Microbial Contamination Screening in Household Drinking Water**

*Suzan Kagan[a,b], Shira Spigelman[c], Sankar Sudhir[a], Thalappil Pradeep[a,d], Hadas Mamane[b*]*

[a] DST Unit of Nanoscience and Thematic Unit of Excellence, Department of Chemistry, IIT Madras, Chennai 600 036, India.

[b] School of Mechanical Engineering, Faculty of Engineering, Tel Aviv University, Tel-Aviv, Israel.

[c] Faculty of Statistics, Tel Aviv University, Tel-Aviv, Israel.

[d] International Centre for Clean Water, 2nd Floor, B-Block, IIT Madras Research Park, Kanagam Road, Taramani, Chennai 600113, India

*Corresponding Author:

Hadas Mamane

School of Mechanical Engineering, Faculty of Engineering

E-mail: hadasmg@tauex.tau.ac.il

**Abstract**

Unsafe drinking water remains a major public health concern globally, particularly in low-resource regions where routine microbiological surveillance is limited. Although *Escherichia coli* is the internationally recognized indicator of fecal contamination, laboratory-based testing is often inaccessible at scale. In this study, we developed and evaluated a two-stage machine-learning framework for predicting *E. coli* presence in decentralized household point-of-use drinking water in Chennai, India using low-cost physicochemical and contextual indicators. The dataset comprised 3,023 samples collected under the People's Water Data initiative; after harmonization, technical cleaning, and outlier screening, 2,207 valid samples were retained. Binary outcomes included total coliform and *E. coli* presence, which were measured using AquaGenx field kits. The predictor variables included physicochemical parameters, water source characteristics, storage and handling practices, and demographic indicators. In the first stage, tree-based classifiers were trained to estimate the probability of the total coliform presence. These out-of-fold predicted probabilities were incorporated as auxiliary features in second-stage models to predict *E. coli* presence.

For model development we used stratified cross-validation, nested validation for early stopping, probability calibration, and threshold selection aligned with public health screening objectives. The selected LightGBM–HistGradientBoosting pipeline achieved cross-validated receiving operator characteristic (ROC)–area under the curve (AUC) values of 0.78–0.85 depending on feature configuration, PR-AUC values > 0.90, and high sensitivity for contaminated samples. Bootstrap hypothesis testing with false discovery rate correction indicated modest but consistent performance improvements relative to the single-stage alternatives.

This framework provides a scalable decision-support tool for prioritizing microbiological testing in resource-constrained environments and addresses an important gap in point-of-use contamination risk assessment.

Beyond predictive modeling, the present study was conducted within an AI-supported field implementation framework that combined student-facing guidance and real-time QC to improve protocol adherence, traceability, and data reliability in decentralized household water monitoring.

# 1. Introduction

Access to safe drinking water is a major global challenge. According to the World Health Organization (WHO), in 2022, an estimated 1.7 billion people worldwide relied on drinking water sources contaminated with fecal matter (World Health Organization). Globally, one in four people still lack access to safe drinking water, highlighting the persistent scale of this public health concern. Such contamination is a major contributor to child mortality, causing approximately 500,000 deaths annually among children under the age of five years due to diarrheal diseases (Geremew and Damtew, 2020). According to WHO guidelines, water for human consumption must be free from disease-causing microbiological agents (World Health Organization, 2017). However, routine regulatory monitoring is commonly based on periodic source-level sampling and may fail to capture microbiological risks at the point-of-use (POU), where water quality may deteriorate during distribution, storage, transport, or household handling. This creates important blind spots in household-level water quality assurance, particularly in decentralized, intermittent, or aging supply systems.

The People's Water Data (PWD) initiative is a structured, quality-controlled household-level monitoring framework that integrates validated field measurements, behavioral surveys, geospatial allocation, and digital reporting. This initiative trains students to collect, test, and interpret water quality information using simple field kits and mobile sensor applications. Importantly, the students undergo 40 h of structured theoretical training. They are not merely data collectors but also act as community ambassadors, a critical component of the initiative's community-based impact model. To date, over 1,600 students have participated in a survey of thousands of households using customized questionnaires and testing kits. Their real-time findings can be shared directly with the residents, enabling evidence-based decisions regarding water use and treatment. This hands-on, student-led approach builds water literacy and empowers communities to make informed decisions on water use and safety. Previous work from this initiative has shown that such decentralized monitoring can generate high-density, field-validated datasets and reveal localized contamination patterns that may remain undetected through conventional centralized surveillance alone.

The PWD has also evolved to include an AI-supported operational layer that strengthens decentralized field implementation. This layer combines two complementary components: a student-facing AI field assistant that provides real-time support during fieldwork, laboratory handling, and digital submission, and a real-time quality control (QC) system that screens incoming records for likely protocol deviations, incomplete execution, and anomalous submission patterns. These tools are not intended to replace supervision or expert validation, but rather to reinforce protocol adherence, improve traceability, and support both individual and group-based learning during practical field sessions.

Khan et al. (2021) developed a machine-learning (ML) model to predict *E. coli* presence in Indian groundwater using readily measurable physicochemical parameters. Their analysis found that turbidity, pH, total dissolved solids (TDS), and electrical conductivity (EC) were the most informative inputs, yielding the best performance for *E. coli* prediction (model $R^2 \approx 0.90$). Similarly, Kuroki et al. (2023) applied support vector machines and related classifiers to municipal tap water data from 26 Nepalese cities, using seven on-site water quality indicators to flag possible *E. coli* contamination. Their approach achieved moderate accuracy (approximately 70% across all cities, which improved to approximately 79% when the outlier case of the capital was excluded) in classifying contaminated samples. These efforts demonstrate that inexpensive physicochemical proxies, such as turbidity, conductivity, and pH, can serve as useful predictors of fecal contamination risk in drinking water, even in data-scarce settings.

Similarly, Ghasemi Tousi et al. (2021) employed ensemble ML methods (random forest and XGBoost) to predict the presence of *E. coli* in irrigation water using only low-cost physicochemical indicators such as turbidity, EC, temperature, total suspended solids (TSS), and rainfall. This approach yielded a high predictive performance (area under the curve (AUC) $\approx$ 0.88; F1 $\approx$ 0.81) and, although it was conducted in an irrigation context, it highlighted the broader relevance of using readily measurable water quality parameters for *E. coli* contamination risk assessment. Similarly, Stocker et al. (2022) employed 12 readily measurable physicochemical proxies in four ML algorithms (Random Forest (RF), SGB, SVM, and k-Nearest Neighbors (KNN)) to predict *E. coli* in agricultural pond water. This approach achieved low prediction error (RMSE 0.24–0.42) and a high coefficient of determination ($R^2 \approx 0.70$), with turbidity and dissolved oxygen (DO) being the most influential predictors.

Hong et al. (2024) developed a ML framework to estimate *E. coli* concentrations in an irrigation pond using both in situ water quality parameters and drone-based RGB imagery. The study tested four algorithms, namely, RF, Gradient Boosting Machine (GBM), Extreme Gradient Boosting (XGB), and KNN, under three input scenarios: water quality only, RGB imagery only, and a combination of both. Their best-performing models (RF and XGB) achieved high test set accuracy ($R^2 \approx 0.93$; RMSE ≈ 3.3–3.9 colony-forming units [CFU]/100 mL) when combining RGB and water quality data. Although conducted in an irrigation context, this study demonstrated the feasibility of integrating low-cost visual imagery with basic water quality measurements to predict microbial contamination, a strategy with potential relevance to drinking water monitoring.

Unlike previous studies that focused on water from the point-of-collection or in agricultural settings, this study targeted water consumed at the household level, where contamination may occur during transport, storage, or handling. This distinction is particularly important because previous work within this manuscript series showed that POU microbial safety is shaped not only by source-water quality or treatment technology, but also by storage conditions, handling practices, infrastructure integrity, maintenance, and user awareness. For example, in household reverse osmosis (RO) systems, microbial contamination was found to persist in treated water despite improvement in several physicochemical parameters, highlighting that treatment alone does not guarantee microbiological safety under real household conditions.

Van der Meulen et al. (2024) conducted a systematic review of predictive models for *E. coli* contamination of urban surface waters, focusing on recreational water bodies within city environments. Out of an initial pool of 305 studies, only 10 met the criteria for providing predictive model specific to the urban context. Most models employ either multiple linear regression (MLR) or ML approaches, such as RF and LightGBM (LGBM), using common input variables, such as turbidity, EC, water temperature, precipitation, and flow rate. The review found that model performance was often limited, with approximately half of the studies reporting $R^2$ values < 0.5. In contrast to the more developed literature on rural and agricultural settings, the authors highlighted a significant knowledge gap regarding urban water systems. They recommended that future studies incorporate seasonal patterns, including known pollution sources (e.g., sewage and stormwater runoff), and consider process-based modeling approaches.

Collectively, these studies demonstrate the potential of low-cost physicochemical indicators for microbial risk prediction. However, other researchers explored complementary approaches that integrate demographic and household contexts. In addition to the physical water metrics, previous studies have also incorporated household and contextual variables to improve predictions.

For example, Ambel et al. (2023), leveraged demographic and socioeconomic features to identify contaminated household water sources in Ethiopia. Using national survey data (with *E. coli* tests on a subset of samples), they trained classifiers that achieved approximately 88% accuracy (AUC of approximately 0.91) in predicting whether a given source contained detectable *E. coli* levels. Notably, their models, which used only non-water variables (e.g., sanitation access, wealth, and geospatial factors), performed nearly as well as those using full water test data, highlighting the value of the household context in contamination risk assessment.

In contrast, our approach fills a critical gap by focusing on POU water quality and leveraging a uniquely comprehensive set of predictors. Although previous studies have addressed water at the source or municipal level, we analyzed the water where it is consumed, that is, after collection, transport, and storage, where contamination risks may increase.

We collected drinking water samples directly from household storage containers and combined them with detailed metadata on storage conditions, handling practices, and on-site treatment behaviors. With this rich, integrated dataset we aimed to develop a context-specific predictive model for microbial contamination that unites low-cost physical measurements with behavioral and environmental factors, a novel contribution to drinking water research in India. Because the PWD framework generates standardized, field-based, household-level data under controlled protocols, it provides a strong foundation for developing practical screening models that can support microbiological prioritization in resource-constrained regions.

Existing predictive studies have largely focused on irrigation water, recreational water, and centralized municipal supplies. Few studies have addressed contamination at the household level POU, where behavioral, infrastructural, and storage-related factors may substantially influence microbial risk. Although the present analysis is specific to Chennai, India, our proposed framework is designed to be transferable to other settings through local calibration.

A plausible physical basis underlies the predictive framework of this study. Turbidity reflects the suspended particles and contamination events that co-occur with microbial risk, whereas conductivity and TDS may capture broader water-quality deterioration and pollution patterns. Oxidation-reduction potential (ORP) and related water chemistry conditions may reflect the redox environments associated with disinfectant residuals or microbial persistence. At the household level, these physicochemical indicators must be interpreted together with storage, handling, and treatment-related variables, because contamination at the POU may arise not only from source water quality but also from recontamination during storage, infrastructure leakage, inadequate maintenance, or unsafe dispensing practices. Thus, our model is not based on a purely statistical association but on the interaction between measurable water quality conditions and known contamination pathways at the household POU.

The main objective of this study was to develop a screening framework that prioritizes confirmatory microbiological testing at the household level POU using low-cost physicochemical and contextual indicators. Rather than replacing microbiological analyses, the model is intended to support field decision-making in settings where routine laboratory testing cannot be applied at a large scale.

This study will contribute to existing literature in three key ways. First, it focuses on POU household drinking water rather than source water, irrigation systems, or centralized supplies. Second, it integrates physicochemical measurements with behavioral, treatment, storage, and demographic variables within a single screening framework. Third, it applies a two-stage modeling strategy in which the probability of total coliform occurrence is leveraged to improve *E. coli* screening performance.

## 2. Materials and Methods

*2.1 Data Collection and Study Design*

Water samples were collected from geographically stratified regions in Chennai, India. Chennai was selected as a relevant real-world testbed because it combines heterogeneous water supply modes, household-level storage dependence, variable treatment behaviors, and infrastructure-related variability, which can strongly influence POU microbial risk. These conditions make it a suitable location for evaluating screening strategies intended for decentralized urban drinking water systems. However, the proposed framework was designed to be adapted to other regions through local recalibration, provided comparable water-quality and household-context data are available.

Routine microbiological testing is expensive and difficult to scale in decentralized household drinking water systems, especially when contamination occurs at the POU after collection or treatment. Therefore, data were obtained from the PWD initiative. Student teams followed standardized protocols and sampled systematically beginning at central landmarks and moved across neighborhoods to ensure spatial diversity. To support protocol adherence during field implementation, the PWD workflow also incorporated an AI-assisted support and monitoring layer. The first component was a student-facing AI field assistant developed around a structured knowledge base of recurrent field questions, covering preparation, sampling, measurements, biological testing, troubleshooting, and digital upload. The second component was a real-time QC layer that screened uploaded records for missing identifiers, duplicate sample IDs, incomplete GPS data, poor GPS accuracy, implausible survey timing, missing photo documentation, logical inconsistencies, and impossible measurement values. These tools were used to support rapid supervisor review and timely feedback during the practical sessions. The surveys captured water source types, including natural sources, such as rivers and wells, municipal supply, bore wells, and public taps. The storage characteristics included container type, material (e.g., metal container), volume, placement relative to the cooking area, and storage duration. Treatment practices included boiling, filtration, reverse osmosis, chlorination, and no treatment. Demographic variables included age, sex, education level, presence of children < 5 years old, and water quality perceptions.

Field measurements included turbidity, TDS, EC, pH, ORP, hardness, and alkalinity. Microbial testing for total coliforms and E. coli was performed using the Aquagenx CBT EC+TC P/A Kit 100-Pack, following the manufacturer's instructions. The outcomes were encoded as binary contamination indicators. Survey records were further subjected to real-time QC screening within the PWD workflow in order to identify likely protocol deviations and prioritize follow-up review.

2.1.1 AI-Supported Field Guidance and Real-Time Quality Control

As part of the broader PWD implementation framework, an AI-supported field guidance and real-time QC layer was developed to improve data reliability and strengthen student performance during decentralized household water surveys. The field-guidance component consisted of a student-facing AI assistant designed to provide immediate support during preparation, site selection, household interaction, field measurements, biological testing, documentation, troubleshooting, and digital upload. The assistant was built around a structured question-and-answer knowledge base reflecting recurrent operational situations encountered by students in the field.

In parallel, a real-time QC system was applied to uploaded records. This system screened submissions across seven domains: record integrity, sample ID traceability, GPS and spatial validity, survey duration, photo completeness, logical consistency, and parameter plausibility. Records were classified into three output categories, namely OK, REVIEW, or ALERT, in order to prioritize follow-up and manual validation. Core QC triggers included missing or duplicate UUIDs, missing sample IDs, GPS accuracy above 30 m, household survey duration below 3 min, water body survey duration below 1 min, incomplete photo sets, temporal anomalies suggestive of batch filling, spatial clustering, and impossible measurement values.

Importantly, this QC layer did not replace expert review. Rather, it functioned as an early-warning mechanism that enabled rapid intervention while field activities were still ongoing. Flagged records were communicated to students within 24–48 h, allowing immediate correction in subsequent submissions. In practice, these flagged cases were used not only for correction at the individual level, but also as part of group-based feedback during practical sessions, thereby supporting shared learning across the cohort. Thus, the AI-supported QC layer contributed simultaneously to data quality assurance and to field-based learning within the PWD methodology.

*2.2 Data Preprocessing and Cleaning*

In addition to post hoc harmonization and technical cleaning, the dataset generation process was supported by a real-time QC workflow embedded within the field implementation process. This enabled likely protocol deviations and incomplete submissions to be flagged during data collection rather than only after export of the full dataset. As a result, part of the data-quality improvement occurred upstream, during field execution itself, through immediate review and feedback. The initial dataset comprised 3,023 samples collected via KoboToolbox, Qualtrics, and Epicollect platforms (665 in Set1 and 2,358 in Set2). Standardization included harmonizing variable encodings, correcting inconsistent spellings, resolving manually entered derivations, and aligning units. Biologically implausible values and duplicate records were removed. RO-treated samples were excluded because such systems cannot introduce new bacterial contamination and would introduce confounding variables into the model.

After technical cleaning, 2,288 valid samples remained, representing 75.6% of the original data. Thereafter, an outlier-based screening procedure removed an additional 81 observations, yielding a final dataset of 2,207 samples, which was approximately 73% of the original entries. An indicator variable denoting the origin of the dataset was included to adjust for potential phase-specific differences.

Physicochemical parameters were evaluated relative to WHO and Bureau of Indian Standards (BIS) standards. For example, the turbidity should ideally remain below 1 NTU, with an acceptable limit of 5 NTU. TDS values exceeding 600 mg/L may become unpalatable, and EC values > 2,500 μS/cm is not recommended. Hardness values > 200 mg/L may contribute to scaling and potential health concerns. Both WHO and BIS require non-detectable total coliforms and *E. coli* in 100 mL of drinking water.

*2.3 Post-Cleaning Distribution of Physicochemical Parameters*

The post-cleaning distributions indicated variability in turbidity, EC, ORP, TDS, hardness, alkalinity, and pH. EC and TDS displayed right-skewed distributions with some high-value outliers, whereas turbidity showed variations consistent with heterogeneous storage conditions. These distributions informed the scaling procedures applied within the cross-validation folds.

*2.4 Statistical Association Between Total Coliforms and E. coli*

Contingency table analysis demonstrated a strong association between total coliform presence and *E. coli* presence. The chi-squared statistic was 366.11 with a p-value < 0.0001, and the estimated odds ratio was 14.28, indicating substantially increased odds of *E. coli* contamination in samples positive for total coliforms. This association motivated the development of a two-stage modeling framework.

*2.5 Modeling Framework*

An 80/20 stratified train–test split was applied, preserving class distributions, given the outcome imbalance (approximately 69.4% positive for *E. coli* and 88.0% for total coliforms).

Stratified five-fold cross-validation was used for model development. Stage 1 involved training tree-based classifiers, including RF, HistGradientBoosting, XGBoost, and LGBM, to predict the total coliform presence. Missing values were handled natively using tree-based algorithms without imputation. Out-of-fold (OOF) predicted probabilities were generated for each observation to avoid information leakage.

Stage 2 incorporated the predicted probability of total coliform presence as an additional feature in the model predicting *E. coli* presence. Probability calibration was applied and classification thresholds were selected to maximize the $F_2$ score, emphasizing recall.

*2.6 Model Evaluation*

Across all the variables, the top-performing configuration was LGBM followed by the two-stage pipeline LGBM, with an AUC value of 0.777934. When the physicochemical variables were excluded, the best configuration was RF followed by LGBM, with an AUC value of 0.769614. Using only physicochemical variables, Random Forest achieved an AUC value of 0.703139. These comparisons demonstrate that contextual and behavioral variables meaningfully improve predictive performance.

Receiver operating characteristic (ROC) analysis showed substantial discrimination above random classification. Shapley Additive Explanations (SHAP)-based feature importance analysis identified latitude, longitude, predicted total coliform probability, turbidity, ORP, TDS, and pH as the most influential features. When physicochemical variables were excluded, contextual variables such as the number of children < 5 years old, storage practices, container placement, and education level emerged as important predictors. Turbidity and EC dominated the model output when only physicochemical variables were used.

A logistic regression model was used to evaluate for interpretability. The coefficients indicate positive associations between turbidity and contamination risk, negative associations with certain pH ranges, and variable effects on ORP and EC. Although logistic regression is generally easy to interpret, its predictive performance in this study was inferior to that of tree-based ensembles.

**3. Results and Discussion**

This study integrated physicochemical, contextual, and behavioral predictors to model microbial contamination at the household level. The two-stage modeling approach leveraged the strong statistical relationship between total coliforms and *E. coli*, using the predicted coliform probability as an informative intermediate feature. This structure improved the stability while preserving the strict separation between the training and validation data.

An additional methodological contribution of this study is that the predictive framework was developed within an AI-supported implementation environment rather than from a static retrospective dataset alone. Within the PWD workflow, AI was used not only downstream for microbial risk prediction, but also upstream to support field execution through student guidance and real-time QC screening. This distinction is important in decentralized monitoring systems because predictive performance depends not only on model architecture, but also on the consistency, traceability, and procedural quality of the field data entering the analytical pipeline. The use of group-based and individual feedback through the QC process also suggests a broader role for AI in field learning. Rather than functioning solely as a computational tool, AI may help build a structured feedback environment in which student monitors improve their field practices while simultaneously generating more reliable scientific datasets. This may be particularly valuable for large-scale student-led and citizen-based environmental monitoring programs.

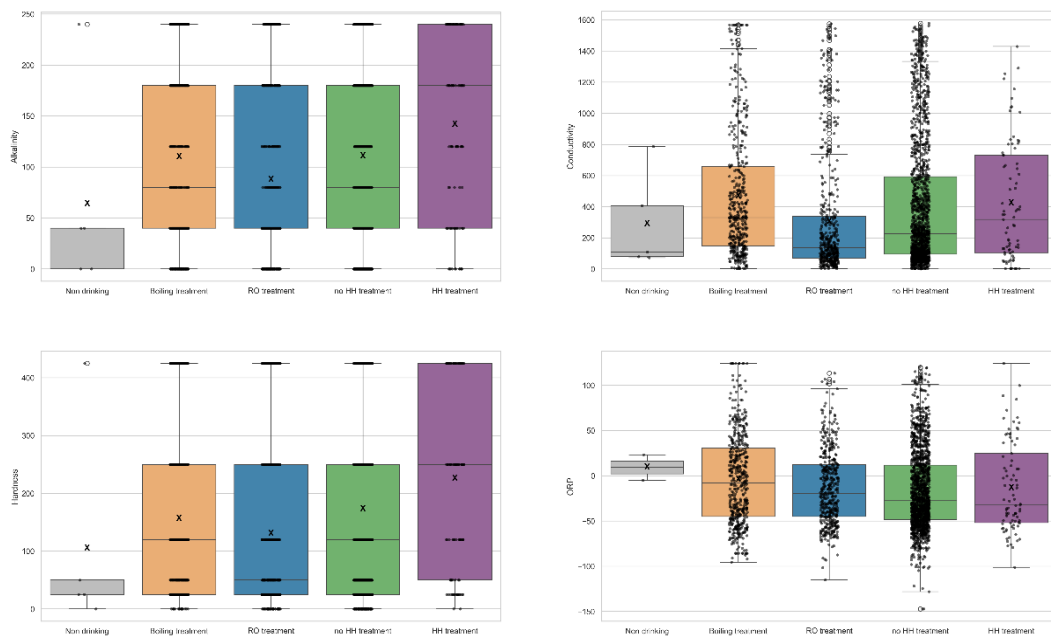

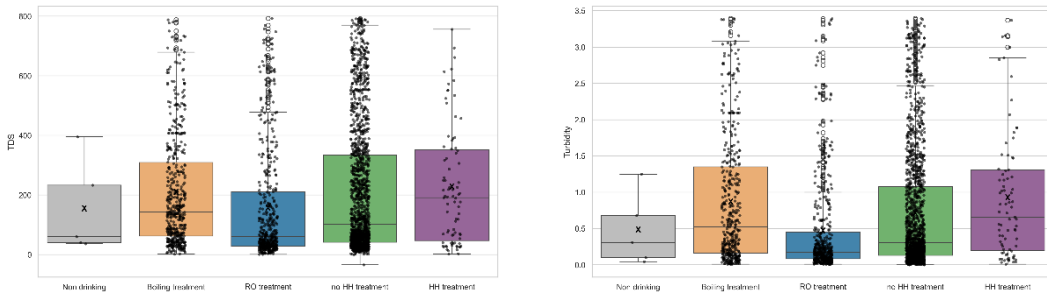

**Figure 1:** Distribution of physicochemical parameters across treatment groups.

The importance of contextual variables highlights the role of storage and handling practices in contamination risk. Unlike irrigation or source-water studies, POU contamination reflects household behaviors and environmental conditions within domestic settings.

This emphasis aligns with public health priorities. In screening contexts, the failure to detect contaminated water may have greater consequences than generating false positives that can trigger confirmatory testing.

After the technical cleaning, the dataset contained 2,285 records. Subsequent outlier-based screening removed 78 additional observations, leaving 2,207 valid samples for analysis. **Figure 2** shows the distributions after the final filtering step.

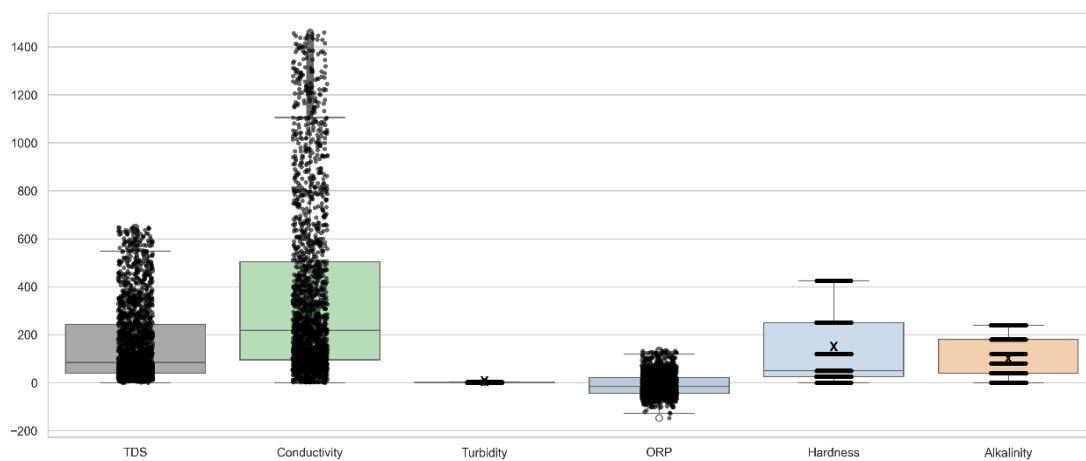

**Figure 2:** Post-cleaning distribution of physicochemical parameters across treatment groups.

To enable meaningful comparisons between variables with different units, all the physicochemical parameters were standardized using z-score normalization. This transformation centered each variable around zero with a standard deviation of one, preserving their relative variation while ensuring a uniform scale across parameters. **Figure 3** shows the distribution of the standardized values across all parameters.

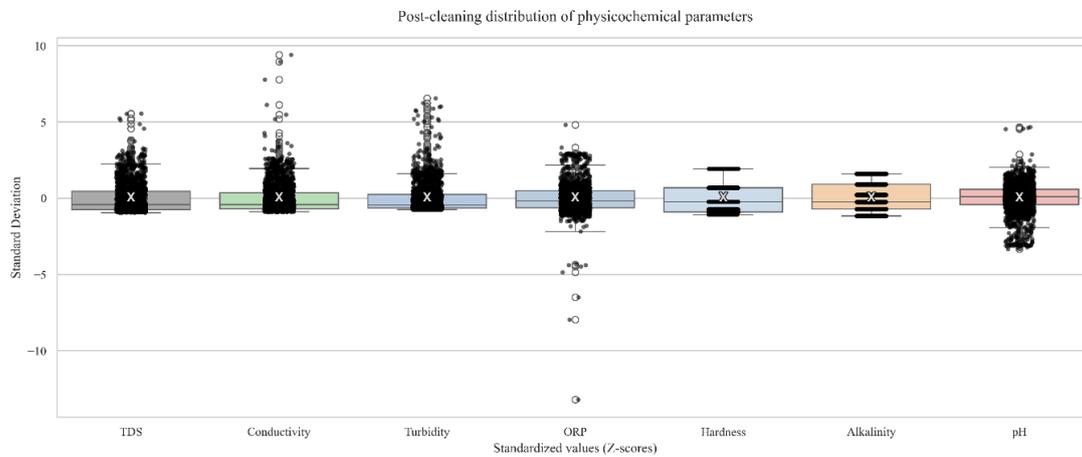

**Figure 3:** Standardized distributions of physicochemical parameters after cleaning and normalization.

*3.1 Preprocessing and Feature Engineering*

The outcome variables were encoded as binary indicators. The predictors included field-measured physicochemical variables and survey-based covariates. One-hot encoding was used for categorical variables with consistent naming across sources. To preserve interpretability and avoid unnecessary transformations, only the physicochemical variables were z-standardized, and this standardization was performed within each training fold (the scaler was fitted on training-only data and applied to validation and fold-held-out tests). Auxiliary Prediction Task Component (PTC) features were not scaled.

*3.2 Two-Stage Approach: Rationale and Architecture*

After data cleaning and preprocessing, we examined whether a two-stage modeling approach was justified in this case. We constructed a contingency table to compare the binary presence of *E. coli* and total coliforms in the dataset. Table 2 presents the proportion of *E. coli* in each coliform group. When total coliforms were absent (value = 0), the rate of *E. coli* was 18.5%. In contrast, when total coliforms were detected (value = 1), the *E. coli* rate increased to 76.4%.

Observations in which *E. coli* was detected in the absence of total coliforms were assumed to be a result of the technical limitations of the testing kit (see Section 4 and supporting information S1).

The chi-squared test indicated a statistically significant association between the two variables ($\chi^2 = 366.11$, $p < 0.0001$), suggesting they are not independent. Moreover, the odds ratio (OR = 14.28) implied that the odds of *E. coli* presence were more than 14 times higher when total coliforms were present.

**Table 1.** Contingency table showing the association between total coliform presence and *E. coli*.

|              | Total coliform = 0 | Total coliform = 1 |
| ------------ | ------------------ | ------------------ |
| *E. coli* = 0 | 216                | 458                |
| *E. coli* = 1 | 49                 | 1,484              |

This provides a strong statistical justification for considering a two-stage modeling approach, where the prediction of *E. coli* is based on the presence of coliforms. This two-stage architecture follows a model-as-feature strategy, in which the prediction of an auxiliary model is used to enrich the input of the main target model.

Consistent with standard practices in data science and applied ML, we split the dataset into 80% for training (1,765 records) and 20% for testing (442 records). The model was developed based entirely on the training set and evaluated on an independent test set to assess its generalizability. Given the imbalanced distribution of the outcome variables (69.4% positive for *E. coli* and 88.0% positive for total coliforms), stratified splitting was used to ensure a balanced representation of the outcome classes across both subsets.

We adopted a two-stage "model-as-feature" framework to predict *E. coli* presence (stage 2) while leveraging the strong statistical association between total coliforms and *E. coli* observed in the data. In stage 1, we trained classifiers to estimate the probability that total coliforms were present; the resulting out-of-fold (OOF) probabilities (denoted as PTC) were then appended to the original feature set as auxiliary predictors and used by stage 2 models to classify *E. coli* presence. The entire procedure is embedded in a stratified five-fold cross-validation (CV) with strict separation between model development and evaluation within each fold to prevent leakage.

3.2.1 Stage 1 (Total coliforms → PTC)

We benchmarked tree-based learners that were robust in tabular and noisy settings: RF, HistGradientBoosting (HistGB), LGBM, XGB, and CatBoost. For each CV fold we split the training portion into an inner training and/validation partitions (85/15, stratified). Early stopping was applied to the inner validation set for boosted models (XGB, LGBM, and CatBoost in 5 × 5 runs) using large iteration budgets and a small learning rate; RF has no early stopping, whereas HistGB uses scikit-learn's built-in early stopping. The tuned model then produces OOF probabilities for the held-out test partitioning of that fold. Concatenating across folds yields PTC for every observation without using its own label during training. We recorded Stage-1 AUC by comparing these OOF probabilities with the true total coliform labels.

3.2.2 Stage 2 (*E. coli*):

For each candidate PTC source (e.g., LGBM, XGB, RF, HistGB, and CatBoost), we augmented the original features with that PTC vector and fit the same family of learners to predict *E. coli*. The same fold logic applies per-fold physicochemical scaling on training only, inner training/validation split, and early stopping on the inner validation data.

Probability calibration was applied only in stage 2. Platt scaling (logistic regression on the model's raw probabilities) was applied to XGBoost and CatBoost, whereas LGBM, RF, and HistGB were calibrated using isotonic regression with a sigmoid fallback if isotonic calibration failed on the validation fold. Within each cross-validation fold we selected the classification threshold that maximizes the Fβ score with β = 2 on the inner validation set (recall-emphasis), and then transferred this F2-optimal threshold to the fold's held-out test split where metrics were calculated. To avoid selection bias, model fitting was nested inside each outer CV fold using only that fold's training data: the training portion was split into an inner training/validation set (stratified), and z-score standardization was fit on the inner train only for physicochemical variables and then applied to the inner-validation and the fold-held-out test.

We used fixed, regularized structural hyperparameters (e.g., max_depth, num_leaves, min_child_samples, subsample, colsample_bytree, and reg_lambda), kept the learning rate within a narrow range, and set a high iteration cap for early stopping during inner validation.

Governs the number of effective trees. Class imbalance was handled per fold (e.g., scale_pos_weight or class_weight) using the prevalences computed on the inner-train only. Models were trained on the inner-train with early stopping against the inner-validation; probability calibration and the $F_2$-optimal threshold (recall-emphasis) were chosen on the inner-validation and then transferred to the fold's held-out test partition. Because class prevalence differed between total coliforms (stage 1) and *E. coli* (stage 2), the imbalance was addressed separately in each stage via per-fold weighting to reduce majority-class bias.

For OOF evaluation, all transformations (standardization), early stopping, calibration, and threshold selection were performed strictly within each fold's training split. Fold-held-out samples were not used to set model parameters, scalers, calibrators, or thresholds. For stage 2 training, the PTC feature was generated OOF, within each outer fold, and stage 1 model was trained only on that fold's training split and used to produce PTC for the corresponding fold-held-out samples. The OOF PTC values were concatenated to form the stage 2 training feature.

*3.3 Evaluation and Reporting*

For each *stage1 → stage2* model pair we constructed OOF predictions of *E. coli* and report the AUC (ROC), PR-AUC (average precision), Brier score (probability reliability), Accuracy,

Precision, Recall, F1, and F2. We also recorded the mean F2-optimal threshold across folds (± standard deviation [SD]), and the mean number of iterations learned by the models.

Additionally, for the selected LGBM→HistGB model we performed Matthew's correlation coefficient (MCC) analysis and calculated specificity to provide balanced-class and false-positive diagnostics. Using calibrated probabilities, we calculated the ROC curve (reported as the ROC-AUC value), and at the chosen $F_2$ threshold we calculated the confusion matrix from which Precision, Recall, F1, and $F_2$ were derived.

Table 3 presents the performance for all possible *stage1→stage2* model combinations. The metrics used calibrated probabilities and a selected $F_2$-optimal threshold per fold. A higher value is better for ROC-AUC/PR-AUC/F1/F2/Recall/Precision/Accuracy, whereas for the Brier score, a lower value is better.

Although ROC-AUC is widely reported, it ranks performance across all thresholds and may appear overly optimistic under a strong class imbalance. Our operational decision used a single threshold and did not prioritize missing positives, thus, models were ranked by the F2 score ($\beta = 2$), which places greater weight on recall than precision. We reported the ROC-AUC and PR-AUC for completeness; however, the F2 value more directly reflects the trade-off relevant to the specific public-health case investigated in this study. This alignment between the metric and objective yielded a more accurate, actionable, and trustworthy model selection.

**Table 2.** Cross-model results (5-fold CV). $P_{TC}$ AUC is the out-of-fold ROC-AUC of Stage-1.

| Rank | Stage1 -> Stage2 | Accuracy | Precision | Recall | F1 | F2 | ROC-AUC | PR-AUC | Brier | F2 Thr (mean) | F2 Thr (SD) | $P_{TC}$ AUC |
|---|---|---|---|---|---|---|---|---|---|---|---|---|
| 1 | LGBM -> HistGB | 0.7575 | 0.7454 | 0.9886 | 0.8499 | 0.9280 | 0.8456 | 0.9180 | 0.1417 | 0.0732 | 0.0592 | 0.7298 |
| 2 | RandomForest -> CatBoost | 0.7700 | 0.7579 | 0.9829 | 0.8558 | 0.9278 | 0.8467 | 0.9234 | 0.1444 | 0.3322 | 0.0346 | 0.7621 |
| 3 | RandomForest -> HistGB | 0.7518 | 0.7405 | 0.9894 | 0.8471 | 0.9271 | 0.8443 | 0.9186 | 0.1423 | 0.0372 | 0.0724 | 0.7621 |
| 4 | XGBoost -> XGBoost | 0.7592 | 0.7477 | 0.9861 | 0.8505 | 0.9270 | 0.8551 | 0.9288 | 0.1412 | 0.2856 | 0.0236 | 0.7247 |
| 5 | HistGB -> LGBM | 0.7569 | 0.7461 | 0.9853 | 0.8492 | 0.9260 | 0.8327 | 0.9063 | 0.1472 | 0.0992 | 0.0625 | 0.7466 |
| 6 | XGBoost -> HistGB | 0.7484 | 0.7384 | 0.9878 | 0.8451 | 0.9253 | 0.8397 | 0.9134 | 0.1455 | 0.0904 | 0.0779 | 0.7247 |
| 7 | HistGB -> HistGB | 0.7484 | 0.7384 | 0.9878 | 0.8451 | 0.9253 | 0.8159 | 0.8955 | 0.1541 | 0.0682 | 0.0580 | 0.7466 |
| 8 | RandomForest -> LGBM | 0.7666 | 0.7563 | 0.9796 | 0.8536 | 0.9250 | 0.8486 | 0.9203 | 0.1416 | 0.0876 | 0.0719 | 0.7621 |
| 9 | LGBM -> XGBoost | 0.7626 | 0.7527 | 0.9804 | 0.8516 | 0.9245 | 0.8558 | 0.9294 | 0.1409 | 0.2974 | 0.0325 | 0.7298 |
| 10 | XGBoost -> CatBoost | 0.7677 | 0.7582 | 0.9772 | 0.8539 | 0.9238 | 0.8419 | 0.9207 | 0.1468 | 0.3582 | 0.0369 | 0.7247 |
| 11 | LGBM -> CatBoost | 0.7615 | 0.7527 | 0.9780 | 0.8507 | 0.9227 | 0.8440 | 0.9219 | 0.1460 | 0.3546 | 0.0384 | 0.7298 |
| 12 | XGBoost -> LGBM | 0.7700 | 0.7615 | 0.9739 | 0.8547 | 0.9224 | 0.8437 | 0.9169 | 0.1429 | 0.0968 | 0.0816 | 0.7247 |
| 13 | LGBM -> LGBM | 0.7637 | 0.7552 | 0.9763 | 0.8517 | 0.9223 | 0.8487 | 0.9179 | 0.1412 | 0.1110 | 0.0590 | 0.7298 |
| 14 | RandomForest -> XGBoost | 0.7824 | 0.7752 | 0.9674 | 0.8607 | 0.9217 | 0.8573 | 0.9300 | 0.1404 | 0.3348 | 0.0321 | 0.7621 |
| 15 | HistGB -> CatBoost | 0.7490 | 0.7415 | 0.9804 | 0.8444 | 0.9211 | 0.8268 | 0.9105 | 0.1534 | 0.3716 | 0.0349 | 0.7466 |
| 16 | HistGB -> XGBoost | 0.7598 | 0.7522 | 0.9755 | 0.8494 | 0.9209 | 0.8378 | 0.9168 | 0.1485 | 0.3334 | 0.0257 | 0.7466 |
| 17 | CatBoost -> LGBM | 0.7280 | 0.7226 | 0.9878 | 0.8346 | 0.9202 | 0.8185 | 0.9035 | 0.1562 | 0.0134 | 0.0248 | 0.7176 |
| 18 | CatBoost -> HistGB | 0.7229 | 0.7182 | 0.9894 | 0.8322 | 0.9199 | 0.8120 | 0.8980 | 0.1577 | 0.0362 | 0.0704 | 0.7176 |
| 19 | CatBoost -> XGBoost | 0.7473 | 0.7413 | 0.9772 | 0.8431 | 0.9187 | 0.8261 | 0.9131 | 0.1535 | 0.3274 | 0.0323 | 0.7176 |
| 20 | HistGB -> RandomForest | 0.7150 | 0.7120 | 0.9902 | 0.8284 | 0.9184 | 0.7957 | 0.8890 | 0.1643 | 0.0316 | 0.0612 | 0.7466 |
| 21 | RandomForest -> RandomForest | 0.7263 | 0.7221 | 0.9853 | 0.8334 | 0.9184 | 0.8068 | 0.8961 | 0.1595 | 0.0678 | 0.0823 | 0.7621 |
| 22 | LGBM -> RandomForest | 0.7286 | 0.7249 | 0.9821 | 0.8341 | 0.9170 | 0.8042 | 0.8957 | 0.1608 | 0.0932 | 0.0757 | 0.7298 |
| 23 | CatBoost -> RandomForest | 0.6997 | 0.7002 | 0.9927 | 0.8212 | 0.9161 | 0.7810 | 0.8817 | 0.1704 | 0.0006 | 0.0005 | 0.7176 |
| 24 | CatBoost -> CatBoost | 0.7405 | 0.7367 | 0.9747 | 0.8392 | 0.9156 | 0.8162 | 0.9094 | 0.1595 | 0.3876 | 0.0364 | 0.7176 |
| 25 | XGBoost -> RandomForest | 0.7246 | 0.7226 | 0.9796 | 0.8317 | 0.9146 | 0.8077 | 0.8983 | 0.1607 | 0.0954 | 0.0772 | 0.7247 |

PTC area under the curve (AUC) is the out-of-fold receiver operating characteristic (ROC)-AUC of Stage-1.

*3.4 Hypothesis Testing Framework and Multiple Testing Problem*

All comparisons used Stage-2 OOF probabilities saved with the exact held-out indices per fold to ensure that each challenger and the reference were evaluated on the same examples (strict pairing).

We used threshold-free ranking metrics to avoid re-using the operating-point targets. For each challenger versus the reference, we tested the hypothesis *H0: Δmetric = 0* vs. *H1: Δmetric ≠ 0* where Δ is challenger minus reference for ROC-AUC and, when available, Average Precision (PR-AUC). We reported the effect sizes, 95% confidence intervals, and adjusted p-values.

Inference was based on a stratified paired bootstrap that resamples within folds, preserving class balance and the pairing structure created by cross-validation. Two-sided p-values were computed from the bootstrap distribution of Δ, and 95% confidence intervals were percentile-based.

For each performance metric, the family comprises challenger-versus-reference comparisons. The false discovery rate (FDR, q-value) was controlled at q = 0.05 using the Benjamini–Hochberg correction, and adjusted p-values or q-values were reported accordingly.

At the fixed global threshold (t∗) learned on training data only, paired 0-1 OOF predictions were compared using McNemar's test. These operating-point tests were treated as a separate family, with FDR adjustment applied within that family.

Random seeds were fixed, Stratified K-Fold was seeded, and XGBoost, LGBM, CatBoost, and HistGB were executed using single-thread deterministic settings. These controls ensured consistent OOF predictions and post-hoc results.

Using the LGBM→HistGB pipeline as the reference, paired-bootstrap analyses indicated mostly small differences in AUC values, with most confidence intervals overlapping zero. Although a few challenger models exhibited positive ΔAUC after FDR adjustment, the observed gains were modest. The LGBM→HistGB pipeline was ultimately selected based on its strong recall at threshold t∗, robust calibration, and stability, while maintaining comparable ranking quality across the leading candidates. **Figure 4** presents the paired-bootstrap ΔAUC effects relative to the reference model using fold-aligned OOF predictions; most intervals overlapped zero and any improvements were modest after FDR adjustment.

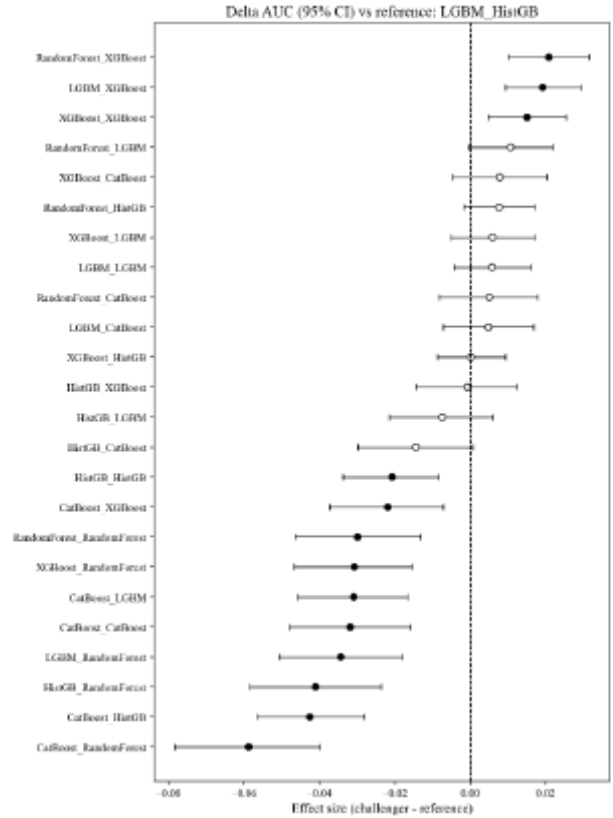

**Figures 4.** Delta AUC (95% CI) vs LGBM→HistGB reference on fold-aligned OOF using stratified paired bootstrap; filled markers q<0.05.

*3.4 Model Selection*

Following the hypothesis-testing results and our recall-first objective, we used the mean OOF F2 (β = 2) score as the primary selection criterion, with ROC-AUC/PR-AUC serving as threshold-free corroboration. Based on this, we selected the two-stage LGBM→HistGB pipeline, as it achieved the highest OOF F2 score with stable per-fold thresholds, and post-hoc paired tests showed no significant AUC/AP advantages for competing models after BH-FDR adjustment. Calibration was maintained via post-fit isotonic calibration.

All the performance metrics below calculated from Stage-2 OOF predictions. In each fold, we (i) trained the inner-train, (ii) calibrated the probabilities on the inner-validation with Platt scaling, (iii) selected the F2-optimal threshold on the validation set, and (iv) evaluated the fold's held-out test. ROC-AUC, PR-AUC, and Brier score used calibrated probabilities, and Accuracy/Precision/Recall/F1/F2 used binary predictions formed by fold-specific F2.

The "with *PTC*" variant augments features with PTC from the Stage-1 LGBM (fit on the full training set for the auxiliary Total_coliform task):

Stage-2 without PTC obtained the following values: ROC-AUC = 0.738, PR-AUC = 0.833, Accuracy = 0.708, Precision = 0.707, Recall = 0.988, F1 = 0.824, F2 = 0.915, and Brier score = 0.178.

Stage-2 with PTC obtained the following values: ROC-AUC = 0.818, PR-AUC = 0.895, Accuracy = 0.770, Precision = 0.757, Recall = 0.985, F1 = 0.856, F2 = 0.929, and Brier score = 0.151.

The auxiliary PTC signal improved both discrimination and calibration; ΔAUC ≈ +0.080 and ΔBrier ≈ −0.027, while maintaining recall (−0.003). Fixed-grid OOF ROC-AUC diagnostics show Stage-1 LGBM peaking at approximately 300–600 estimators with a mild taper beyond ~1,000, and Stage-2 HistGB (no PTC) degrading steadily as max_iter grows (100 → 2,000), consistent with overfitting risk. With PTC, HistGB attains its best AUC already near ~100 iterations and then plateaus (≈ 0.85) through higher complexity, indicating that a compact model suffices and validating the use of early stopping in Stage-2. Notably, the F2 threshold was far more stable across folds in the presence of PTC (mean 0.290; SD 0.028) than without PTC (mean, 0.277; SD 0.141).

*3.5 Explainability*

To interpret the models, we used SHAP, which attributes the marginal contribution of each feature to each feature its marginal contribution to an individual prediction. In the beeswarm plots, each dot is one sample; the color encodes the feature value (blue = low, pink/red = high) and the horizontal position shows whether the feature pushes the prediction toward higher or lower contamination risk.

Stage-1 (LGBM): The main drivers are physicochemical indicators (TDS_ppm, pH, Conductivity_S, ORP_mV, Turbidity_AVG_NT) - with several household and demographic variables (e.g., Number_of_children_under_the_age_of_5, container-handling practices) that play secondary roles. These directions are intuitive: poorer water-quality measurements and riskier handling behaviors increase the predicted contamination, whereas protective practices reduce it.

Stage-2 without PTC (HistGB): In the absence of PTC, the model leans more heavily on spatial cues (latitude/longitude) along with the same physicochemical signals and a subset of household-practice variables. This suggests location acts as a proxy when the model is "hungry" for explanatory signal.

Stage-2 with PTC (HistGB). Adding PTC markedly reshaped the explanation: PTC became the dominant contributor, and the remaining features (including latitude/longitude and physicochemical set) receded to supporting roles. This aligns with the idea that thestage-1 Total-coliform probability captures abroad contamination risk that is highly informative for *E. coli.*

The bar chart of the mean |SHAP| across the three setups confirms this pattern quantitatively: physicochemical factors dominate Stage-1; spatial variables become prominent only without PTC; once PTC is included, it accounts for the largest average contribution, whereas while other features form a secondary tier. Together, the beeswarm plots (direction and sample-level effects) and the mean-|SHAP| bars (aggregate contributions) show that *PTC* improves the accuracy and simplifies the feature-importance structure, reinforcing the value of the two-stage design.

We also conducted a subgroup analysis of the fully labeled set (n = 2,207), stratified by reported household water-treatment practices, using the fixed deployment threshold t* = 0.294. Table 4 indicates that the model maintained a high recall across the major subgroups, with $F_2$ scores of 0.969 for the 'HH' treatment, 0.956 for the 'boiling' treatment, and 0.928 for the 'No HH' treatment. Two very small groups, namely, the RO treatment group (n = 6) and the others (n = 5), yielded unstable estimates and were not interpreted.

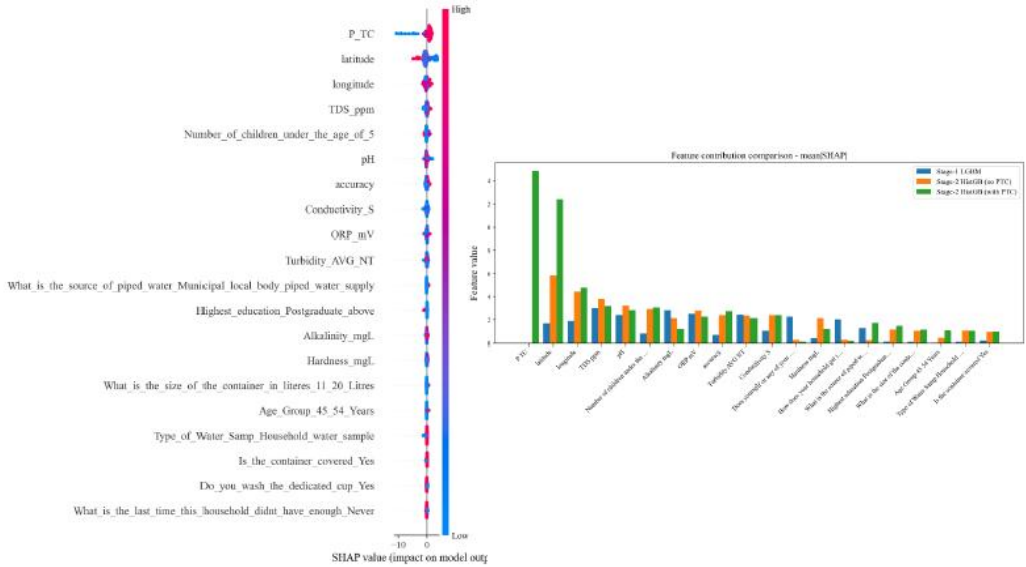

**Figures 5.** SHAP plots (left to right): Stage-1; Stage-2 without *PTC*.

**Table 3.** Subgroup performance by treatment type (full dataset).

| How_did_you_treat_mapped | n | Precision | Recall | F1 | F2 | Specificity |
|---|---|---|---|---|---|---|
| Other | 5 | 1.0000 | 1.0000 | 1.0000 | 1.0000 | 1.0000 |
| HH treatment | 92 | 0.9130 | 0.9844 | 0.9474 | 0.9692 | 0.7857 |
| Boiling treatment | 479 | 0.9479 | 0.9584 | 0.9532 | 0.9563 | 0.8390 |
| RO treatment | 6 | 0.8000 | 1.0000 | 0.8889 | 0.9524 | 0.5000 |
| no HH treatment | 1625 | 0.9258 | 0.9283 | 0.9271 | 0.9278 | 0.8432 |

*3.6 Test Set Evaluation*

For the final test evaluation, stage-1 is refit the full training data to generate PTC for the test set only, and stage-2 is trained on the full training data using the OOF PTC feature. This preserves the strict separation between training and evaluation and avoids stacking leakage.

To determine the true generalization power of the two-stage pipeline, we applied it to a held-out test dataset (442 records). Using the global F2-optimal threshold learned on the full training set ($t^* \approx 0.481$), the LGBM→HistGB pipeline achieved the following test values: ROC-AUC

= 0.768, PR-AUC = 0.872, Accuracy = 0.740, Precision = 0.758, Recall = 0.919, F1 = 0.831, F2 = 0.881, and Brier score = 0.174.

The decision threshold for the test was markedly higher than the very low per-fold CV thresholds. This was expected because small CV splits tend to push the F2 optimum toward low thresholds to maximize recall, whereas a single global threshold learned from the full training distribution balances recall and precision more conservatively. In practice, the test threshold of approximately 0.48 maintains a very high recall (0.919) with a notable improvement in precision relative to permissive CV thresholds.

Compared with similar studies, the model demonstrates strengths in class-imbalance-aware metrics, with PR-AUC (~0.87) and recall (~0.92) indicating competitive performance for microbial water contamination prediction under real-world imbalance. The use of F2 and a calibrated Brier score (0.174) further strengthens our evaluation, as many studies focus primarily on accuracy or ROC-AUC and do not evaluate PR-AUC or calibration. ROC-AUC (0.768) was moderate, overall accuracy was good but not exceptional.

**Figure 6** shows how performance varies with the decision threshold on the held-out test set. Recall decreases as the threshold increases, precision increases, and both F1 and F2 scores peak at intermediate thresholds. We included the F1 curve to provide a complete view of the trade-off. The dashed line indicate the global F2-optimal threshold (t* ≈ 0.481), which was selected to prioritize high recall while maintaining precision at an acceptable level.

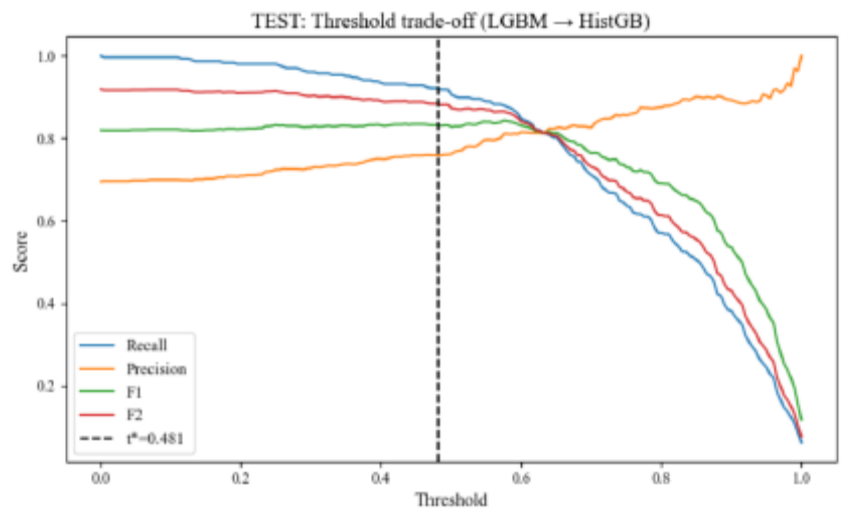

**Figure 6.** Threshold-performance trade-off on the test set.

For practical deployment, our framework can be used by trained field teams or local monitoring programs to identify households requiring priority microbiological confirmation. Rather than testing all samples in the laboratory, practitioners could first collect low-cost field measurements and household-context data, apply the screening model, and then conduct a direct confirmatory microbial analysis of the highest risk cases. This would be particularly valuable in decentralized settings where laboratory throughput, cost, and response time constrain routine monitoring.

Although calibrated using data from Chennai, India, the framework is not limited to this region and can be adapted to other decentralized drinking-water systems using locally collected data, contextual variables, and site-specific threshold calibration.

## 4. Limitations and Challenges

This study had several limitations. First, the dataset is specific to Chennai, and external validation is required before it can be applied elsewhere. Second, some variables known to affect microbial persistence, including free residual chlorine and water temperature, were not available in the dataset. Third, field-based microbial test kits introduce practical detection constraints relative to laboratory reference methods. Finally, although the model supports screening and prioritization, it is not intended to replace confirmatory microbiological analyses. The AI-supported field assistant and real-time QC layer were not only conceptual components of the PWD framework, but were also deployed in real time during field implementation across more than 400 survey submissions. However, their independent quantitative contribution to downstream predictive model performance was not isolated as a separate analytical objective in the present study. Future work should evaluate this contribution more explicitly while also assessing their effect on upstream data quality and field learning.

## 5. Future Directions

The remaining analysis steps included a finalized model evaluation of the held-out test set and expanded hypothesis testing. Long-term directions include feature engineering, integration into mobile applications for field officers, and the evaluation of transferability to other regions beyond India.

In addition, future studies should validate our framework across diverse geographic settings and incorporate additional variables influencing microbial persistence, such as free residual chlorine and water temperature. Further work should also compare field-based measurements with laboratory methods and evaluate real-world implementation to support its use as a screening and prioritization tool.

Future work should extend not only the predictive modeling component, but also the AI-supported operational layer of the PWD framework. This includes expanding the student-facing field assistant, embedding real-time QC outputs directly into field dashboards, and evaluating how AI-supported feedback influences protocol adherence, student behavior, and field learning over time. More broadly, this line of work may contribute to the development of AI-enabled field learning analytics for decentralized environmental monitoring systems.

## 6. Conclusion

In this study, we designed a recall-prioritized two-stage ML framework. We showed that this framework can effectively screen for *E. coli* contamination in household POU drinking water. Moreover, by integrating physicochemical measurements with contextual household variables, this model provides a scalable decision-support tool for prioritizing laboratory testing in resource-constrained environments.

**Acknowledgements**